%% file: main.tex
\documentclass[a4paper,twoside]{article}

\usepackage{epsfig}
\usepackage{subcaption}
\usepackage{calc}
\usepackage{amssymb}
\usepackage{amstext}
\usepackage{amsmath}
\usepackage{amsthm}
\usepackage{multicol}
\usepackage{pslatex}
\usepackage{apalike}
\usepackage{algorithm2e}
\usepackage[bottom]{footmisc}

\usepackage{float}
\usepackage{multirow}
\usepackage[table]{xcolor}
\usepackage{comment} 
\usepackage[unicode=true]{hyperref} 
\usepackage{adjustbox}
\usepackage{cuted}
\usepackage{capt-of}

\usepackage{SCITEPRESS}     
\usepackage{orcidlink}

\definecolor{gold}{rgb}{1.0, 0.6, 0.6} 
\definecolor{silver}{rgb}{1.0, 0.8, 0.6} 
\definecolor{bronze}{rgb} {1.0, 1.0, 0.6} 

\definecolor{darkest}{rgb}{0.6, 0.6, 0.6} 
\definecolor{darker}{rgb}{0.8, 0.8, 0.8} 
\definecolor{dark}{rgb} {0.9, 0.9, 0.9} 

\begin{document}

\title{Improving Adaptive Density Control  for 3D Gaussian Splatting}

\author{\authorname{Glenn Grubert*\sup{1}\sup{2}\orcidlink{0009-0007-9423-7533}, 
Florian Barthel*\sup{1}\sup{2}\orcidlink{0009-0004-7264-1672}, 
Anna Hilsmann\sup{2}\orcidlink{0000-0002-2086-0951} 
and Peter Eisert\sup{1}\sup{2}\orcidlink{0000-0001-8378-4805}
}
\affiliation{* equal contributions}
\affiliation{\sup{1}Humboldt Universit\"at zu Berlin, Berlin, Germany}
\affiliation{\sup{2}Fraunhofer HHI, Berlin, Germany}
\email{glenn.grubert@student.hu-berlin.de}
}


\input{chapters/0_abstract}

\onecolumn \maketitle \normalsize \setcounter{footnote}{0} \vfill

\input{chapters/1_introduction}

\input{chapters/3_preliminaries}
\input{chapters/2_related_work}

\input{chapters/4_method}

\input{chapters/5_experiments}

\input{chapters/6_conclusion}

\bibliographystyle{apalike}
{\small\bibliography{main}}

\input{chapters/7_appendix}

\end{document}

%% file: chapters/0_abstract.tex
\abstract{
3D Gaussian Splatting (3DGS) has become one of the most influential works in the past year. Due to its efficient and high-quality novel view synthesis capabilities, it has been widely adopted in many research fields and applications. Nevertheless, 3DGS still faces challenges to properly manage the number of Gaussian primitives that are used during scene reconstruction. Following the adaptive density control (ADC) mechanism of 3D Gaussian Splatting, new Gaussians in under-reconstructed regions are created, while Gaussians that do not contribute to the rendering quality are pruned. We observe that those criteria for densifying and pruning Gaussians can sometimes lead to worse rendering by introducing artifacts. We especially observe under-reconstructed background or overfitted foreground regions. To encounter both problems, we propose three new improvements to the adaptive density control mechanism. Those include a correction for the scene extent calculation that does not only rely on camera positions, an exponentially ascending gradient threshold to improve training convergence, and significance-aware pruning strategy to avoid background artifacts. With these adaptions, we show that the rendering quality improves while using the same number of Gaussians primitives. Furthermore, with our improvements, the training converges considerably faster, allowing for more than twice as fast training times while yielding better quality than 3DGS. Finally, our contributions are easily compatible with most existing derivative works of 3DGS making them relevant for future works.
}

%% file: chapters/1_introduction.tex
\section{\uppercase{Introduction}}
\label{sec:introduction}

In the past year, 3D Gaussian Splatting (3DGS) \cite{first_gs} has emerged as a powerful tool for real-time 3D scene reconstruction. By representing a scene with a large collection of 3D Gaussian primitives (or splats), each being described by a position, a color, a density, a scale and a rotation, 3DGS is able to model complex structures with high fidelity while allowing very fast rendering. 

\begin{figure}[!ht]
\centering
\includegraphics[width=\linewidth]{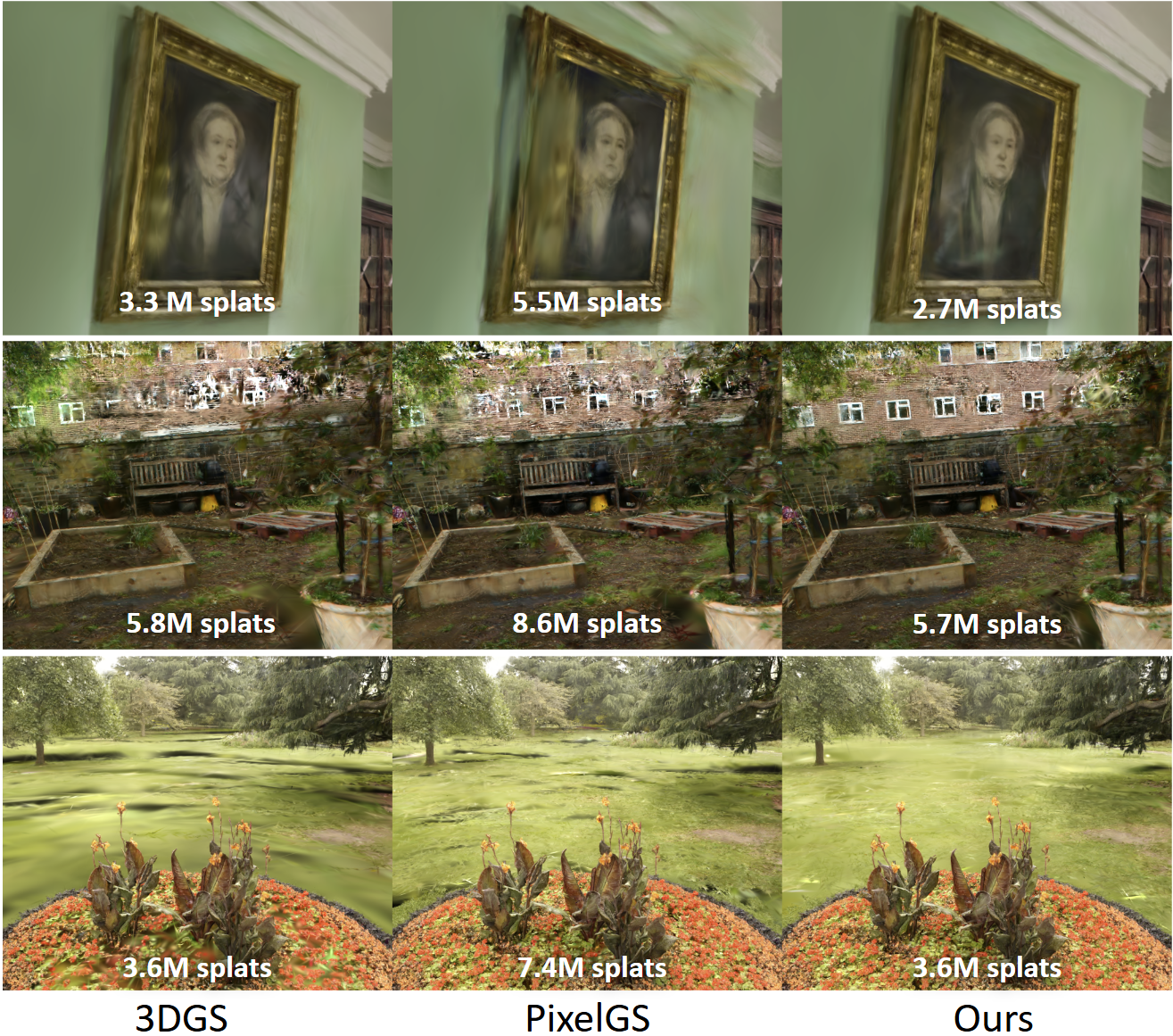}
\caption{Rendered novel views from the drjohnson, garden and flowers scene using 3DGS, PixelGS and our method. Our method produces better backgrounds, creates less artifacts and uses the same number of Gaussians as 3DGS. }
\label{fig:teaser}
\end{figure}

Although rendering using splats has been around for many years \cite{ewa_splatting,botsch,pfister,ren}, 3DGS, for the first time, allows differential rendering. This makes the method applicable for a large range of research fields, such as gradient-based novel-view synthesis, generative models, style transfer, scene editing, animation, and many other \cite{luiten2023dynamic,liu2023stylegaussian,Barthel_2024_CVPR,3DGSzip2024}. In its main application, the novel view synthesis, a 3D scene is generated from multiple 2D views of a target environment along with their respective camera positions. Specifically, with the differentiable rendering property of 3DGS, a loss between a rendered view and a target view can be computed, and the Gaussian attributes in the scene are updated via gradient descent. This is done until the optimization converges and a resulting 3D scene is constructed, which can be viewed from arbitrary novel views. Before, this was achieved mainly using neural radiance field methods, e.g.\ NeRF, Mip-NeRF, Instant NGP or Plenoxels \cite{mildenhall2020nerf,barron2021mipnerf,plenoxels,mueller2022instant}. Although NeRF-based approaches produce very high rendering quality for novel views, they are slow to train and slow to render. This is because they have to query a small neural network multiple times to render a single pixel. Additionally, NeRFs store a 3D scene implicitly in the weights of a neural network, making it difficult to incorporate them into explicit 3D environments, such as game engines or virtual reality settings. 3DGS, in contrast, represents the scene with explicit 3D Gaussian splats, enabling a smooth integration into explicit 3D environments.

Despite its success, 3DGS encounters certain limitations. One major challenge lies in managing in the number of Gaussian primitives and their distribution within . scene For example, in coarse and smooth regions, a small number of large Gaussians can be sufficient, whereas regions with fine and complex structures require a large number of small Gaussians. 3DGS addresses this through an adaptive density control (ADC) mechanism, where the Gaussians are cloned or split .if their associated position changes frequently during optimization and deleted if their opacity drops below a certain threshold. Although the ADC algorithm of 3DGS already yields high-quality results in many cases, it tends to generate too many Gaussians in simple regions, leading to overfitting and, conversely, under-reconstructs highly complex regions, such as grass. In this paper, we propose an improved ADC algorithm designed to enhance reconstruction quality while reducing the number of Gaussian primitives required, surpassing the efficiency of state-of-the-art methods through the following contributions:

\begin{enumerate}
    \item \textbf{Exponential Gradient Thresholding}: We introduce an exponentially ascending gradient threshold that accelerates convergence and improves stability during optimization, particularly in complex regions.
    \item \textbf{Corrected Scene Extent for Adaptive Cloning and Splitting}: By adjusting the scene extent  calculation to account for view depth, our approach ensures a more accurate distribution of Gaussians, optimizing both dense and sparse regions.
    \item \textbf{Significance-Aware Pruning}: Our novel pruning criterion evaluates the impact of each Gaussian on overall scene fidelity, reducing unnecessary primitives while preserving high reconstruction quality.
\end{enumerate}

In the following sections, we will first provide a detailed overview of 3DGS and its ADC algorithm (section \ref{sec:preliminaries}). We then discuss related work on the ADC algorithm of Gaussian Splatting (section \ref{sec:related_work}). Section \ref{sec:method} describes our proposed method, followed by experiments and an ablation study (section \ref{sec:experiments}) and a conclusion (section \ref{sec:conclusion}).

%% file: chapters/3_preliminaries.tex
\section{\uppercase{Preliminaries}}
\label{sec:preliminaries}

\subsection{Gaussian Primitives}
\label{subsec:gaussians}

3D Gaussian splatting uses 3D Gaussian distributions \( F: \mathbb{R}^3 \to \mathbb{R}^+ \) to represent a scene \cite{first_gs}. A 3D Gaussian function is parameterized by a \( (3\times3) \) covariance matrix \( \Sigma \) and a 3D mean vector \( \mu \). \( \Sigma \) describes the extent of a Gaussian function in 3D space, while \( \mu \) characterizes its position or center \cite{ewa_splatting}. The covariance \( \Sigma \) can be parameterized by a 4D quaternion vector \( q \), describing the rotation, and a 3D scaling vector \( s \). The kernel \( K \) of a multivariate Gaussian distribution is defined as follows \cite{error_based_dens}:

\begin{equation}
    K[\mu, \Sigma](x) = \exp \left( -\frac{1}{2} (x - \mu)^\top \Sigma^{-1} (x - \mu) \right) 
    \label{eq:kernel}
\end{equation}

To obtain a Gaussian primitive, a 3D Gaussian function is extended by two additional attributes: an opacity value \( o \) and a feature vector \( c \). The opacity describes how transparent a primitive is displayed, and \( c \) modulates the color in which a primitive is displayed \cite{first_gs}.

For simplicity, from now on these primitives will be referred to as Gaussians. Thus, a Gaussian \( g_k \) can be described as a 4-tuple \( (\mu_k, \Sigma_k, o_k, c_k) \). A 3D Gaussian model scene representation \( G \) can be described as a set of \( n \) Gaussians \( \{g_1, \dots, g_n\} \).

\subsection{Alpha Blending}
\label{subsec:alpha_blending}
The first step towards rendering is to select only those Gaussians \( \Omega \subseteq G \) that appear within the camera view. 

For rendering, these 3D Gaussians are projected onto the 2D image plane of a selected view. The obtained 2D Gaussians \( g_k' = (\mu_k',\Sigma_k',o_k,c_k) \), from now on referred to as splats, then contribute to the color of the pixels they cover on that plane \cite{first_gs}. \( \mu_k' \) and \( \Sigma_k' \) are the view-dependent parameterizations of the 2D projection, and \( o_k \) is the opacity, while \( c_k \) is the feature vector inherited from \( g_k \).

Let \( L = (g_1',\dots,g_m') \) be a list of ascending depth-sorted splats that cover a pixel \( p \). \( K[\mu, \Sigma] \) denotes a Gaussian kernel defined in Equation \ref{eq:kernel}. Alpha blending can then be described as follows \cite{error_based_dens}: 

\begin{flalign}
    color_p &= \sum_{k=1}^{m} c_k \cdot \overbrace{\alpha_k \cdot t_k}^{w_k}
    \label{eq:color_sum} \\
    \alpha_k &= o_k \cdot K[\mu_k', \Sigma_k'] (p) 
    \label{eq:transparency} \\
    t_k &= \prod_{j=1}^{k-1} (1 - \alpha_j)
    \label{eq:transmittance}
\end{flalign}

Equation \ref{eq:color_sum} iterates over all \( m \) splats in the list and computes the final RGB \( \text{color}_p \) for a pixel \( p \). Therefore, the color \( c_k \) and the alpha-blending coefficient \( w_k \) are multiplied per splat. \( w_k \) weights the color contribution of a splat to a pixel. Equation \ref{eq:transparency} computes the alpha value for a splat, given a pixel \( p \). The computation of \( w_k \) considers the opacity and density of a given splat (transparency \( \alpha_k \)) and of all its predecessors (transmittance \( t_k \), Eq.\ \ref{eq:transmittance}). Thus, the frontmost splat's color impact is weighted more strongly than that of the following splats. In the implementation by \cite{first_gs}, the algorithm halts once a specific alpha saturation is reached, rendering the contribution of subsequent splats to the pixel color negligible.

\subsection{Adaptive Density Control}
\label{subsec:adc}
Densification aims to increase the number of Gaussians in the scene to better match its structure. Often, the Gaussians obtained from initialization are too sparse to allow for detailed scene reconstruction. Densification compensates for this by cloning and splitting specific Gaussians, thereby increasing the density and total number of Gaussians in the scene. This process is performed during each \textit{densification interval}, which is set by default to every 1000 steps until until the training reaches 15k iterations.

While densifying the scene plays an essential role for improving the rendering quality, pruning is equally important to maintain a a manageable number of Gaussian primitives. Ideally, only Gaussians that significantly enhance the scene's rendering quality are retained. In 3DGS, this is achieved by removing Gaussians with low opacity or by removing Gaussians or those excessively large relative to the scene's dimensions. Pruning is performed directly after the densification step. 

Additionally, every 3000 steps an opacity reset is performed, reducing the opacity of all Gaussians. This, combined with the regular pruning step, allows the training algorithm to identify and remove Gaussians that are no longer necessary, ensuring an efficient representation of the scene.


Algorithm \ref{alg:densify_and_prune} describes the densification and pruning procedures as proposed by \cite{first_gs}. Here, \( \tau_k \) denotes the average 2D positional gradient magnitude, \( o_k \) the opacity, and \( \max(s_k) \) the maximum scaling factor of a Gaussian \( g_k \). \( P_\text{dense} \) denotes the \textit{percent dense} hyperparameter and \( T_\text{grad} \) the \textit{gradient threshold} hyperparameter. \( e_\text{scene} \) is the precomputed \textit{scene extent} and \( o_\text{min} \) the predetermined \textit{minimum opacity}. 

\begin{algorithm}[!ht]
 \caption{Densify and Prune Algorithm from 3DGS}
 \KwData{Scene of Gaussians $G$}
 \label{alg:densify_and_prune}
  \For{$g_k \in G$}
    {
        \CommentSty{//Densification}
        
        \If{$\tau_k \geq T_\text{grad}$}
        {
            \eIf{$\max(s_k) > P_\text{dense} \cdot e_\text{scene}$}{splitGaussian($g_k$);}{cloneGaussian($g_k$);}
        }
        \CommentSty{//Opacity Pruning}
        
        \If{$o_k < o_\text{min}$}
        {
            pruneGaussian($g_k$);
        }
        \CommentSty{//Size Pruning}
        
        \If{$\max(s_k) > 0.1 \cdot e_\text{scene}$}
        {
            pruneGaussian($g_k$);
        }
    }
\end{algorithm}

\subsubsection*{Cloning and Splitting}
\label{subsubsec:cl&sp}
The cloning method creates an exact copy of the selected Gaussian and adds it to the scene. In contrast, the splitting algorithm replaces a Gaussian with a fixed number (\( N_\text{children} = 2 \) by default) of child Gaussians. These split child Gaussians are positioned within the deleted parent Gaussian's extent. For sampling, the algorithm uses the probability distribution modulated by each Gaussian. The children's scaling is the parent Gaussian's scaling, downscaled by the factor \( \frac{5}{4} \cdot \frac{1}{N_\text{children}} \). All other properties are inherited unchanged from their parent Gaussian. 

Recently added Gaussians evolve differently from their parent Gaussians, as they have no prior Adam \cite{adam_optim} optimization momentum.

\subsubsection*{2D Positional Gradient}
\label{subsubsec:2d_grad}
The 2D positional gradient magnitude of a Gaussian \( g_k \) is averaged over all \( N_\text{views} \) views between two densification steps. Since the average concerns different 2D projections (splats) of the same Gaussian for different views, \( g_k^v = (\mu_k^v, \Sigma_k^v, o_k, c_k) \) denotes the splat of \( g_k \) for view \( v \). Based on this, \( \tau_k \) can be defined as follows \cite{pixelgs}:

\begin{equation}
    \tau_k = \frac{1}{N_\text{views}} \sum_{v = 1}^{N_\text{views}} \left\| \frac{\partial \mathcal{L}(I_v,I'_v) }{\partial \mu_k^v} \right\|_2 
\label{eq:2d_grad}
\end{equation}

Here, \( \mathcal{L} \) denotes the image loss between the ground truth image \( I_v \) and the rendered image \( I'_v \) corresponding to view \( v \).

%% file: chapters/2_related_work.tex
\section{\uppercase{Related Work}}
\label{sec:related_work}

Although Gaussian Splatting surpasses state-of-the-art reconstruction methods in terms of quality, significant potential remains for further improvement. A wide range of research has proposed various enhancements and adaptations for 3DGS. In the following, we will give a brief overview of recent methods that have made improvements to the 3DGS adaptive density control mechanism. These methods are either motivated by reducing the number of Gaussian primitives without losing rendering quality or by improving rendering quality while using a comparable or slightly increased number of Gaussians. To the first type of methods we will refer to as \textit{compaction methods} and to the second we will refer to as \textit{quality improvement methods}. Our proposed method can be classified as the latter.

\vspace{0.5cm}
\noindent{\textbf{Compaction methods}} focus primarily on pruning Gaussians with minimal impact on the rendering quality. For instance, \textit{LightGaussian} \cite{fan2023lightgaussian} uses a knowledge distillation post-processing step that inputs a trained 3DGS scene and outputs a reduced 3DGS scene by eliminating Gaussians that do not meet a global significance threshold. Afterwards, the scene is optimized again (without densification) to address minor inaccuracies introduced by the pruning process. \textit{Color-Cued Efficient Densification} \cite{Kim_2024_CVPR}, on the other hand adapts the criteria that decides wether a Gaussian will be densified. Instead of only using the positional 2D gradient, as in 3DGS, \cite{Kim_2024_CVPR} additionally uses the gradient of the color attributes. This enhancement leads to the generation of a significantly reduced number of Gaussians. Furthermore, \cite{Kim_2024_CVPR} learns a binary mask as an additional 3DGS attribute that determines whether a Gaussian will be rendered. Similarly, \textit{mini splatting} \cite{mini-splatting} assigns an importance attribute to each Gaussian representing the probability of it being sampled. Moreover, \cite{mini-splatting} prunes Gaussians that have insufficient intersection with rays during the rendering of the training views.

\vspace{0.5cm}
\noindent{\textbf{Quality improvement methods}} on the other hand, aim to improve the rendering quality without focusing on a reduction in Gaussian primitives. For example, \textit{Revising Densification in Gaussian Splatting} \cite{error_based_dens} introduces an error based densification criteria that focuses on densifying regions where the structure is different to the target image. This is done by calculating a structural similarity (SSIM) loss for each pixel, and assigning it to the Gaussians that have contributed to the color of the respective pixel proportionally. Consequently, Gaussians with a high loss value are split or cloned during densification. Moreover, \cite{error_based_dens} adapts the opacity initialization for newly cloned Gaussians. In 3D Gaussian Splatting, the newly cloned child inherits its parent Gaussian's opacity. This decision decreases the impact of any Gaussian $g_k$ subsequent in the alpha blending process, as the cloned child and parent Gaussian together have a greater impact on the transmittance $t_k$ than the parent Gaussian alone had before cloning. To counteract this bias, they equally split the opacity between the parent Gaussian and its clone child in a manner that keeps $t_k$ before and after cloning approximately the same.
\begin{equation}
    o_{\text{new}} = 1-\sqrt{1-o_{\textit{old}}} 
    \label{eq:op_after_cloning}
\end{equation}
Here, $o_{\text{old}}$ is the parent Gaussian's opacity before cloning, and $o_{\text{new}}$ is the parent’s and clone child's opacity after cloning.

Similar to \cite{error_based_dens}, \textit{PixelGS} \cite{pixelgs} proposes an adapted densification criteria. Specifically, they modify the gradient-based densification from 3DGS so that it is in relation to the number of pixels that each Gaussian influences. By simply dividing the gradient with the number of pixels that render the respective Gaussian, large Gaussians in background regions are densified less frequently, even though they might have a high positional gradient. To further encourage densification in background regions while reducing it in foreground regions, they scaled the 2D gradient \( \tau_k \) of a Gaussian \( g_k \) based on its distance from the camera. These improvements yield a refined pixel-aware and depth-scaled \( \tau_k \) defined as follows:

\begin{flalign}
    \tau_k' &= \frac{\sum_{v = 1}^{N_\text{views}} f_k^v \cdot \text{count}(g_k^v) \cdot \left\| \frac{\partial \mathcal{L}(I_v, I'_v)}{\partial \mu_k^v} \right\|_2}{\sum_{v = 1}^{N_\text{views}} \text{count}(g_k^v)}
    \label{eq:corrected_2d_grad}
\end{flalign}

\begin{flalign}
    f_k^v &= \text{clip}\left( \frac{\text{depth}(g_k^v)}{\gamma \cdot e_\text{scene}}, [0,1] \right)
    \label{eq:depth_scaling}
\end{flalign}

Here, \textit{count} and \textit{depth} denote functions that compute the pixel count and the image depth of a rendered splat, respectively. \( \gamma \) is a selectable hyperparameter ($0.37$ by default), and \( f_k^v \) is the depth-scaling factor. With this procedure, \textit{PixelGS} achieves state-of-the-art rendering quality, however, at the cost of considerably more Gaussian primitives. With our proposed method, we exceed the rendering quality of \textit{PixelGS} and \textit{Revising Densification in Gaussian Splatting} while using a similar number of Gaussians as in 3DGS.

%% file: chapters/4_method.tex
\section{\uppercase{Method}}
\label{sec:method}
In the following three sections we will introduce our contributions to improve the adaptive density control of 3DGS. As our proposed components are highly compatible with existing methods, we will use a combination of current state-of-the-art methods as our basis and build our additions on top. Specifically, we will use the pixel weighted gradient based densification from PixelGS \cite{pixelgs}, as formulated in Equation \ref{eq:corrected_2d_grad} and combine it with the adapted opacity for cloned Gaussians from \cite{error_based_dens} as described in Equation \ref{eq:op_after_cloning}.

\subsection{Corrected Scene-Extent}
\label{subsubsec:corrected_extent}
In 3D Gaussian Splatting, the \textit{scene extent} (\( e_\text{scene} \)) is used to determine whether a Gaussian that satisfies the \textit{gradient threshold} condition should be cloned or split. A larger \( e_\text{scene} \) favors cloning, while a smaller \( e_\text{scene} \) favors splitting. Furthermore, \( e_\text{scene} \) impacts size pruning, as the size of Gaussians is evaluated relative to the \textit{scene extent}. In the default implementation of 3DGS, the value of \( e_\text{scene} \) is computed based on the camera positions as follows:

\begin{flalign}
e_\text{scene} &= 1.1 \cdot \max_{i = 1}^{N_{\text{cam}}} \left\| \overline{C} - C_i \right\|_2
\label{eq:baseline_extent} \\
\overline{C} &= \frac{1}{N_\text{cam}} \sum_{i = 1}^{N_\text{cam}} C_i
\label{eq:camera_center} 
\end{flalign}

\begin{figure}[t]
\centering
\includegraphics[width=\linewidth]{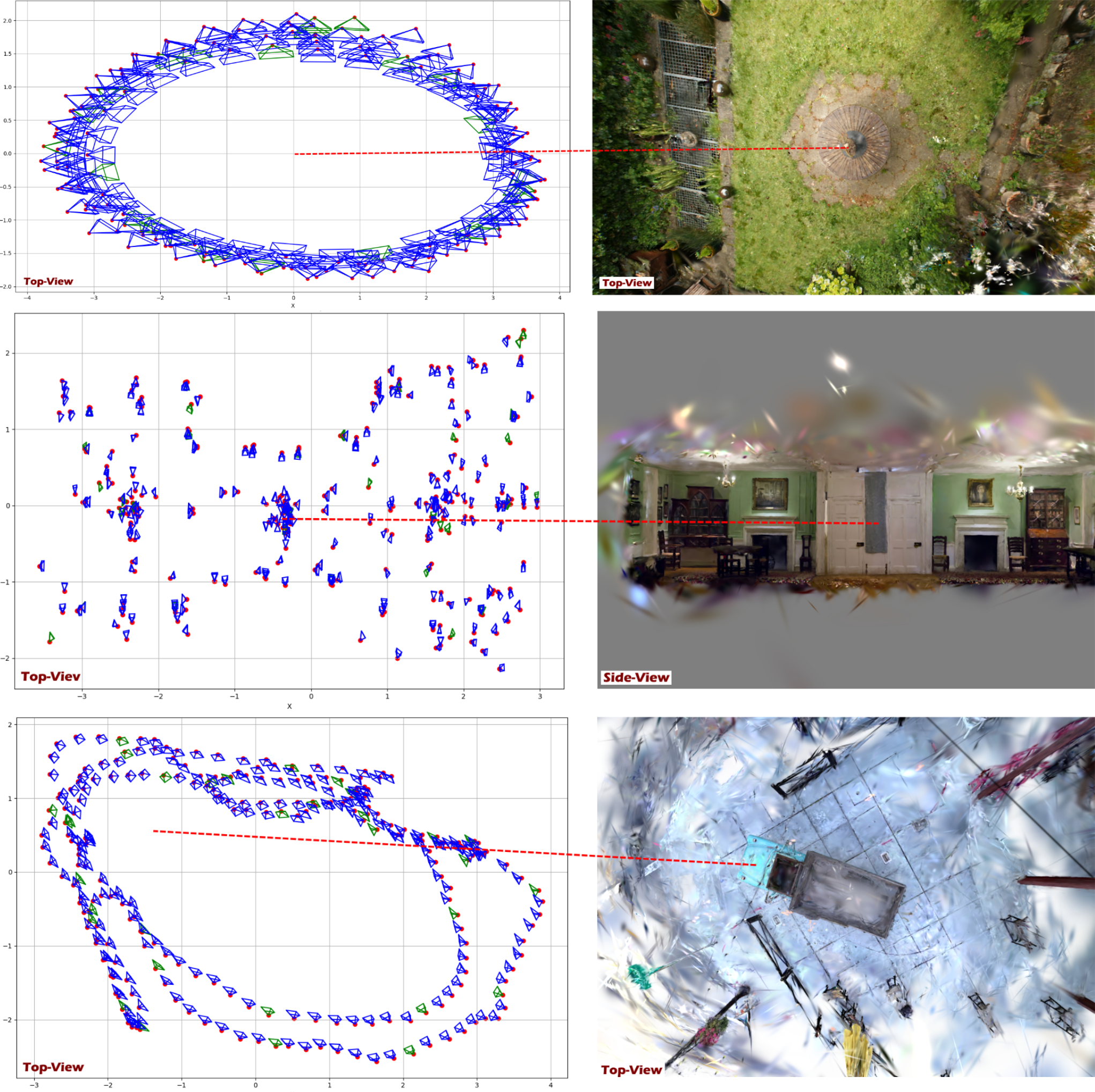}
\caption{Camera locations of the Garden scene, the drjohnson scene and the truck scene. Here, the camera positions are closely distributed around the table in the center of the garden.}
\label{fig:garden_cameras}
\end{figure}

Here, \( N_\text{cam} \) denotes the number of camera views, with \( C_i \) representing the position of the \( i \)-th camera. With this formulation, the \textit{scene extent} is proportional to the distance of the furthest camera location in relation to the average camera position. This makes the optimization algorithm heavily dependent on the capturing process of the scene. If, for example, the camera is orbiting around a small object within a large environment, the \textit{scene extent} can be very small, although the scene is very large. An example for this is shown in Figure \ref{fig:garden_cameras} with the Garden scene of MipNeRF360. Here all of the cameras are close to the center, leading to a very small \textit{scene extent}.


Subsequently a proportionally large outdoor scene like Garden obtains a smaller \textit{scene extent} than a proportionally small indoor scene like Drjohnson, where cameras are distributed throughout the entire volume. 
The \textit{scene extent} so far only reflects the distance of camera positions but not the actual scene volume. During training, this bias will lead to some Gaussians being split or pruned, even though they might be sized appropriately for such a big scene. 

To encounter this, we propose a corrected scene, that is not dependent on the camera locations, but instead dependent on the SfM point cloud that is used as an initialization for 3DGS. Specifically, we formulate the new \textit{scene extent} as follows: 

\begin{equation}
    e_\text{scene}' = \frac{1}{N_{\text{SfM}}} \sum_{i = 1}^{N_{\text{SfM}}} \left\| \overline{C} - p_i \right\|_2 
    \label{eq:corrected_extent} 
\end{equation}

Here $N_\text{SfM}$ is the number of SfM points $p$, and $\overline{C}$ is the averaged camera position from Equation \ref{eq:camera_center}. 

Overall, this correction provides a \textit{scene extent} of the same order of magnitude, better corresponding to the naturally perceived extent for each scene. 

\subsection{Exponentially Ascending Gradient Threshold}
\label{subsubsec:asc_grad_thresh}

\begin{figure}[t]
\centering
\includegraphics[width=\linewidth]{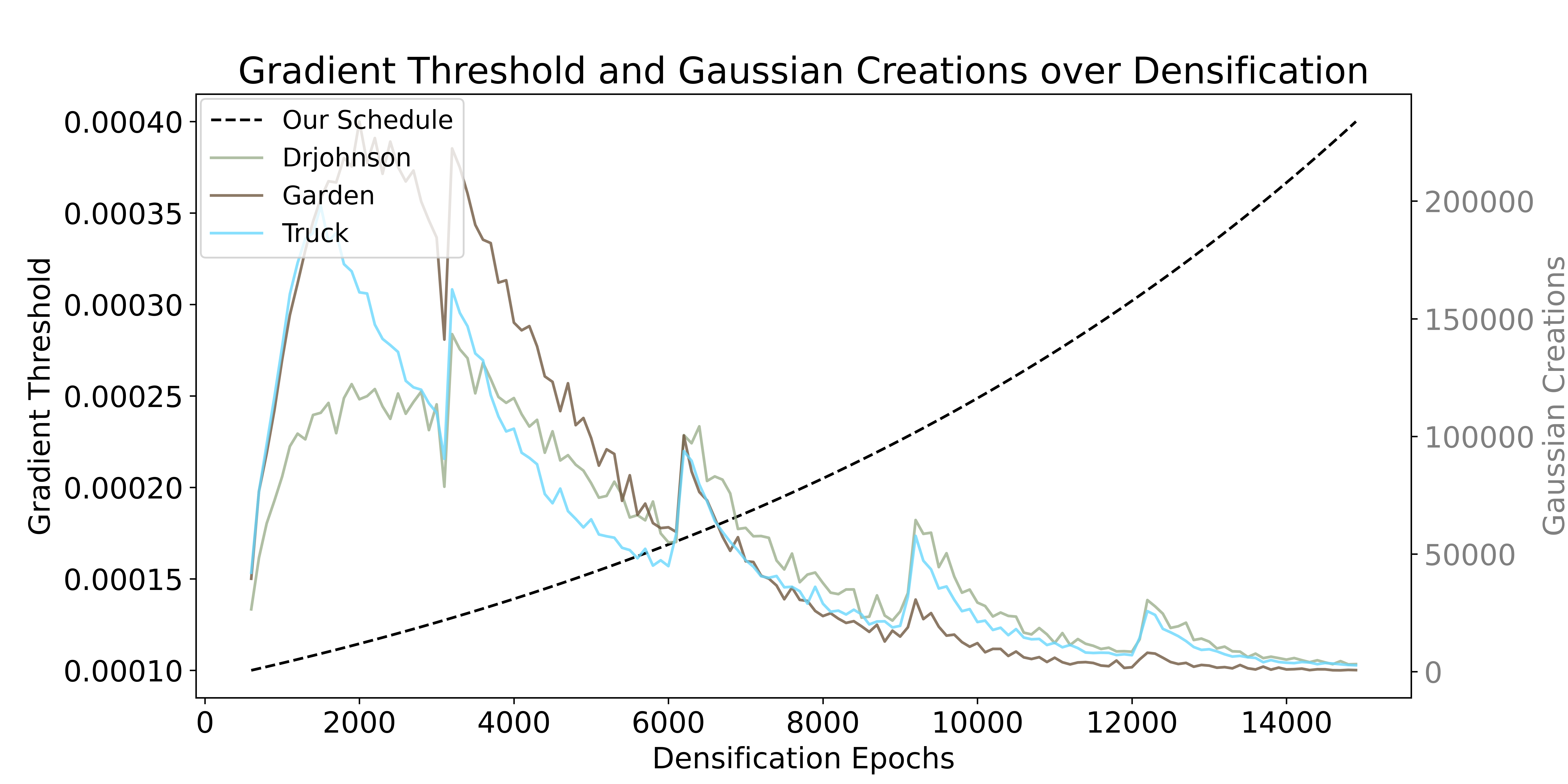} 
\caption{Our proposed \textit{gradient threshold} schedule over the densification interval compared against the number of Gaussians created by our model. The number of densified Gaussians clearly corresponds to our exponential schedule.}
\label{fig:exp_schedule}
\end{figure}

To control the number of Gaussians split or cloned during densification, 3DGS compares the accumulated positional gradient of each splat against a predefined threshold. Gaussians with high gradients, indicating frequent movement during optimization, are selected for densification. Intuitively, this approach assumes that if a Gaussian moves frequently, it is effectively filling multiple places at once. Splitting or cloning these Gaussians distributes the work on two separate Gaussians.


In 3DGS, this threshold is set to the fixed value of 0.0002. We argue that this setting is not ideal for fast convergence. Specifically, during early optimization steps, only few Gaussians exist in the scene, hindering convergence without spawning new Gaussians. Vice versa, at the end of the training, the scene is already constructed of many Gaussians, where additional densification can lead to overfitting. To encounter both problems, we propose an ascending \textit{gradient threshold} that starts at a low value of 0.0001, allowing many Gaussians to densify during the beginning, and ends at a high value of 0.0004, where only those Gaussians with a very high positional gradient are cloned and split. Figure \ref{fig:exp_schedule} visualizes that schedule and its implications for the densification process. 
The exponential scheduling for the threshold $T_i$ in iteration $i$ can be described as follows:

\begin{equation}
    T_i = \exp \left(\ln(T_{\text{s}}) \cdot (1 - \frac{i}{i_{\text{max}}}) + \ln(T_{\text{f}}) \cdot \frac{i}{i_{\text{max}}}\right)
    \label{eq:thresh_scheduling} 
\end{equation}

\begin{figure}[t]
\centering
\includegraphics[width=\linewidth]{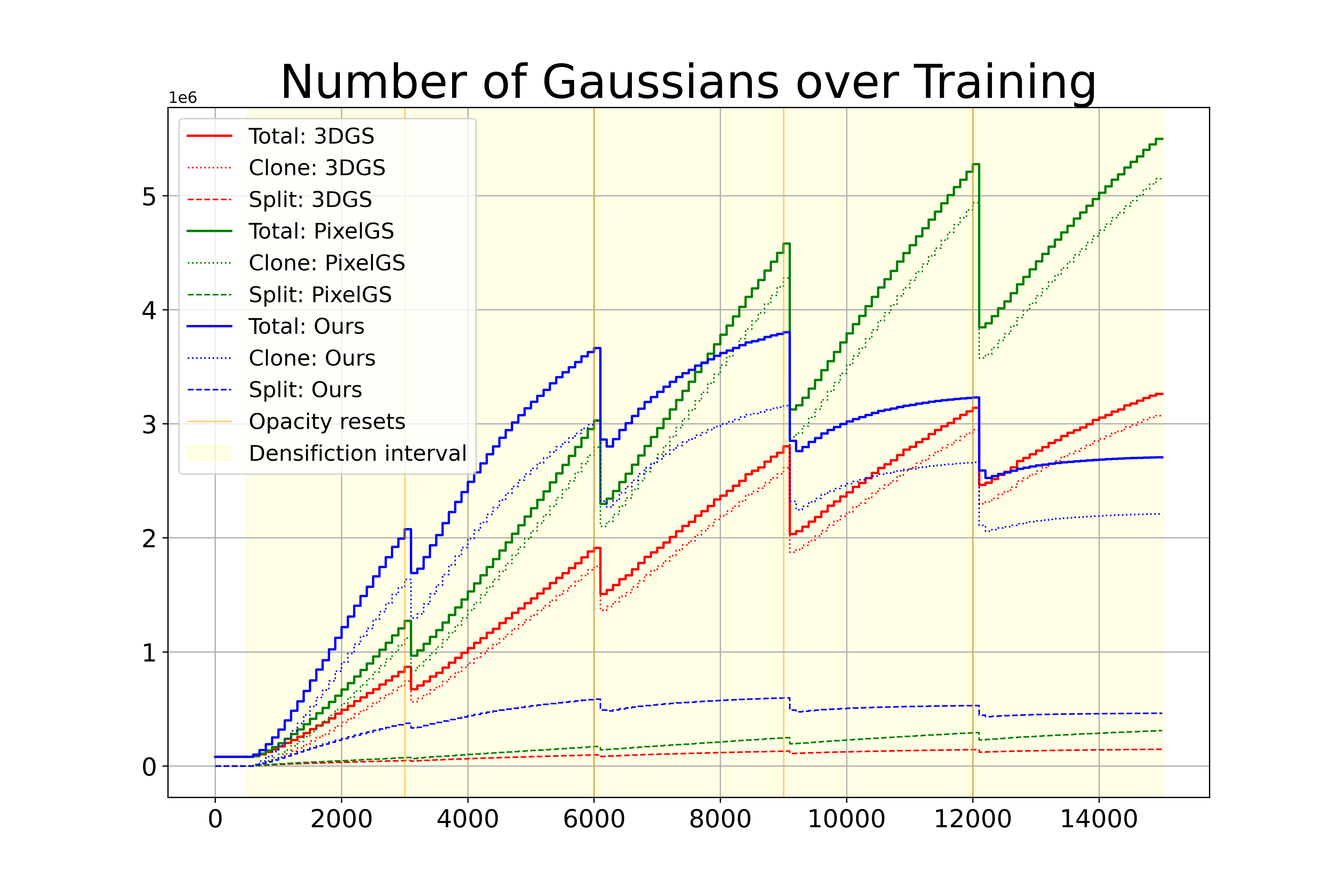} 
\caption{Number of Gaussians during the training. With our proposed exponentially ascending \textit{gradient threshold}, we produce many Gaussians at the beginning of the training and few at the end. Here, we trained with the Drjohnson scene.}
\label{fig:number}
\end{figure}

Here $i_\text{max}$ denotes the number of iterations, $T_s$ is the initial, and $T_f$ the final value for the \textit{gradient threshold}. As shown in Figure \ref{fig:number}, with this configuration, the number of Gaussians rises much quicker, compared to 3DGS and PixelGS, during early training steps, while it declines towards the end. After 15000, the densification stops and our method produces a similar number of Gaussians as 3DGS.

\subsection{Significance-aware Pruning}
\label{subsubsec:new_pruning}

\begin{figure}[t]
\centering
\includegraphics[width=\linewidth]{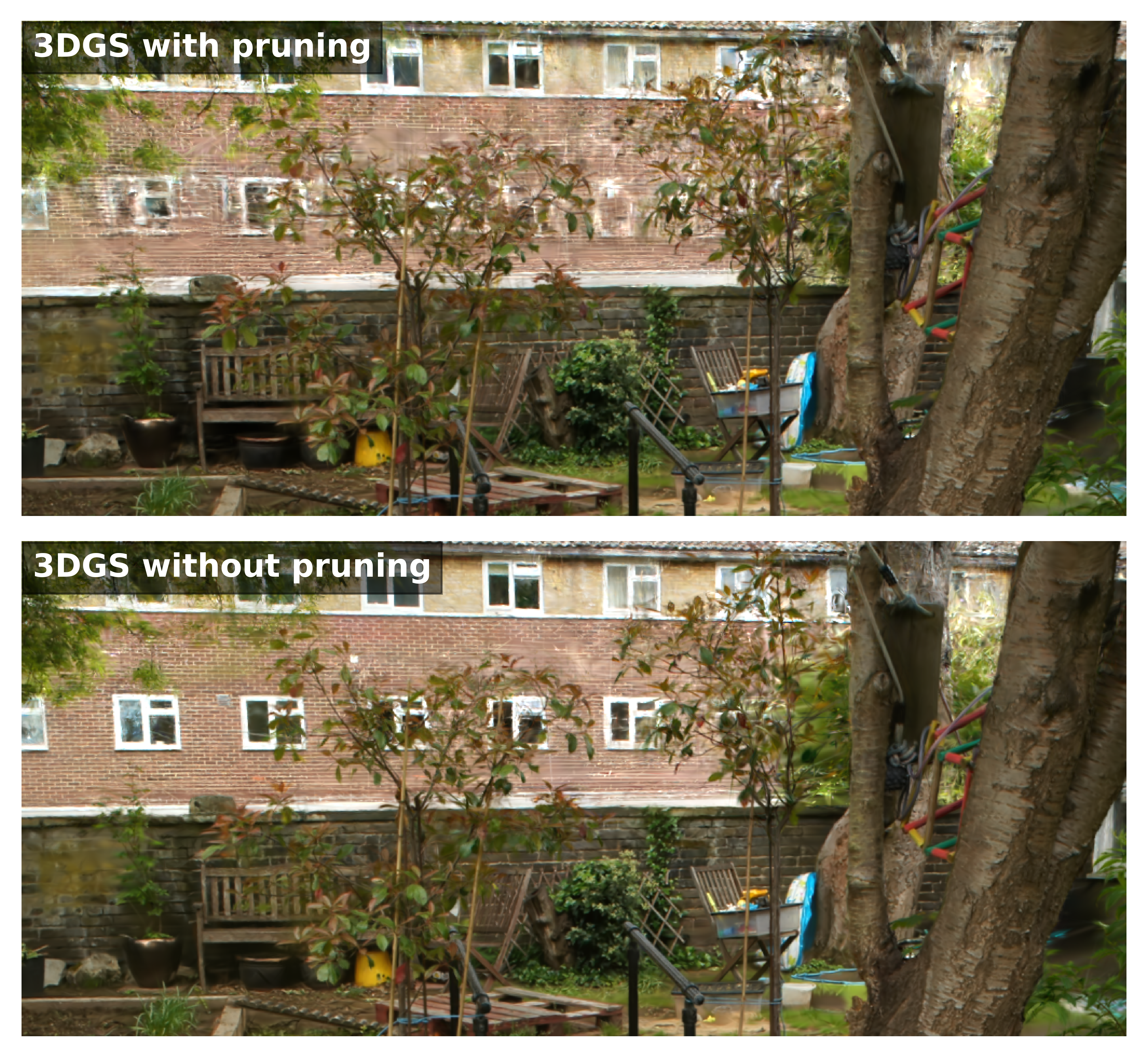} 
\caption{Selected novel view from the Garden scene trained with the baseline (3DGS) in two configurations, one with and the other without pruning.}
\label{fig:pruning_causes_trouble}
\end{figure}

In 3D Gaussian Splatting, the primary objective of pruning lies in reducing the overhead caused by storing and processing unnecessary Gaussians that do not effectively contribute to scene reconstruction. However, finding those specific Gaussians remains a challenge. Unless the opacity or the scale of a Gaussian is set to zero, it is not trivial whether the Gaussian can be removed without harming the rendering quality. Figure \ref{fig:pruning_causes_trouble} indicates that the baseline pruning algorithm performs pruning decisions conflicting with scene reconstruction.

An effective pruning algorithm therefore should minimize the chance of pruning Gaussians that are essential for scene reconstruction. We propose an improved pruning strategy that effectively balances the trade-off between reconstruction quality and scene compactness by more accurately considering a Gaussians contribution to the scene.

The pruning algorithm in 3DGS consists of size and opacity pruning. We, however, observe that size pruning has an adverse effect on both, scene reconstruction and scene compactness. By removing a large Gaussian from the scene, multiple small Gaussians end up replacing it, thus increasing the overall number of Gaussians in the scene. Moreover, a large Gaussian with high opacity essentially contributes to scene reconstruction. Therefore, simply removing them most likely damages existing scene structures.

On the other, Opacity pruning only takes a Gaussian's opacity $o_k$ into account. However, this approach does not fully reflect a Gaussian's contribution to the scene. If a Gaussian $g_k$ has a large extent, it has a more significant impact on alpha blending and $\text{color}_p$, especially if it contributes with high transmittance $t_k$. Therefore, the indicator that best reflects a Gaussian's contribution is the alpha blending coefficient $w_k$.

To account for this discrepancy our proposed pruning algorithm integrates both, a Gaussians opacity $o_k$ and a Gaussians alpha blending coefficient $w_k$ into the pruning decision. 
Our pruning method first selects all $N_\text{prune}$ Gaussians with opacities below $o_\text{min}$. From this selection, it then prunes only those Gaussians with accumulated $w_k$ values within the bottom $N_\text{prune}$.

\begin{flalign}
    \sigma_k &= \sum_{p=1}^{N_\text{pixel}} w_k^p 
    \label{eq:accum_weights} \\
    \Sigma &= \{\sigma_1, \cdots, \sigma_n \}
    \label{eq:sigma_set}
\end{flalign}

This accumulation $\sigma_k$ is computed over all $N_\text{pixel}$ pixels of all views between two densification steps. This approach spares Gaussians with low opacity but high significance to the scene reconstruction.

Over all, by considering both opacity $o_k$ and alpha blending coefficients $w_k$ in the pruning decision, our approach evaluates a Gaussians contribution to the scene more precisely. This precise evaluation enables performing reliable pruning decisions throughout scene training.
Hence our method more effectively optimizes the tradeoff between reconstruction quality and scene compactness.

\begin{algorithm}[!ht]
 \caption{Our revised Densify and Prune Algorithm with the corrected \textit{scene extent} $e_\text{scene}'$, the exponential ascending \textit{gradient threshold} $\T_i$ and the significance pruning.}
 \KwData{Scene of Gaussians $G$}
 \label{alg:our_densify_and_prune}
  \For{$g_k \in G$}
    {
        \CommentSty{//Densification}
        
        \If{$\tau_k' \geq T_i$}
        {
            \eIf{$\max(s_k) > P_\text{dense} \cdot e_\text{scene}'$}{splitGaussian($g_k$);}{cloneGaussian($g_k$);}
        }
        \CommentSty{//Significance Pruning}
        
        \If{$o_k < o_\text{min}$}
        {
            \If{$\sigma_k \in \text{bottom}(N_\text{prune},\Sigma)$}
            {
                pruneGaussian($g_k$);
            }
            
        }

    }
\end{algorithm}

%% file: chapters/5_experiments.tex
\begin{table*}
\caption{Metrics and number of Gaussians averaged over all scenes from, MipNeRF 360 \cite{barron2021mipnerf}, Tanks and Temples \cite{t&t} and Deep Blending \cite{deep-blending} datasets respectively. We retrieved metrics for Plenoxels \cite{plenoxels}, INGP \cite{mueller2022instant}, and MipNeRF \cite{barron2021mipnerf} from \cite{first_gs}. The LPIPS score for Revising Densification \cite{error_based_dens} is missing, since they use a different LPIPS calculation that is not publicly available.}
\label{tab:combined_metrics}
\resizebox{\linewidth}{!}{
\begin{tabular}{|l|rrrr|rrrr|rrrr|}

\hline
Dataset & \multicolumn{4}{c|}{Mip-NeRF 360} & \multicolumn{4}{c|}{Tanks and Temples} & \multicolumn{4}{c|}{Deep Blending} \\
Method / Metric & \scriptsize PSNR$\uparrow$ & \scriptsize SSIM$\uparrow$ & \scriptsize LPIPS$\downarrow$ & \scriptsize \#Gaussians$\downarrow$ 
& \scriptsize PSNR$\uparrow$ & \scriptsize SSIM$\uparrow$ & \scriptsize LPIPS$\downarrow$ & \scriptsize \#Gaussians$\downarrow$ 
& \scriptsize PSNR$\uparrow$ & \scriptsize SSIM$\uparrow$ & \scriptsize LPIPS$\downarrow$ & \scriptsize \#Gaussians$\downarrow$ \\

\hline
Plenoxels   &   23.08   &   0.626   &   0.463   &   &   21.08   &   0.719   &   0.379   &   &   23.06   &   0.895   &   0.51    &\\
INGP-Big    &   25.59   &   0.699   &   0.331   &   &   21.92   &   0.745   &   0.305   &   &   24.96   &   0.817   &   0.390   &\\
MipNeRF     &   27.69   &   0.792   &   0.237   &   &   22.22   &   0.759   &   0.257   &   &   29.40   &   0.901   &   0.245   &\\

\hline
3DGS      &	27.39	&	0.813	&	0.218	&   \textbf{3.3M}  &	23.74	&	0.846	&	0.178	&	\textbf{1.8M}&	29.50	&	0.899	&	0.247	&	2.8M \\

PixelGS \cite{pixelgs}   &	27.53	&	0.822	&	\textbf{0.191}	&	5.5M &	23.83	&	0.854	&	0.151	&	4.5M &	28.95	&	0.892	&	0.250	&	4.6M \\

Revising Dens. \cite{error_based_dens}   &	27.61	&	0.822	&		    &	\textbf{3.3M}&	23.93	&	0.853	&	&	\textbf{1.8M}  &	29.50	&	\textbf{0.904}	&		    &	2.8M \\

Ours       &	\textbf{27.71}	&	\textbf{0.824}	&	0.193	&	\textbf{3.3M} &	\textbf{24.18}	&	\textbf{0.860}	&	\textbf{0.149}	&	2.4M  &	\textbf{29.63}	&	0.902	&	\textbf{0.240}	&	\textbf{2.3M} \\
\hline
\end{tabular}
}
\end{table*}

\section{\uppercase{Experiments}}
\label{sec:experiments}

Combining these efforts yields Algorithm \ref{alg:our_densify_and_prune}, which we implemented into our model. For computing pixel-aware and depth-scaled 2D gradients we rely on implementations provided by \cite{pixelgs}, without modifying any predetermined parameters. The Git repository containing all adaptations is available at \url{https://github.com/fraunhoferhhi/Improving-ADC-3DGS}.

For qualitative and quantitative evaluation, we used the 3DGS evaluation pipeline provided by \cite{first_gs}. Our method was tested on all real-world scenes shared by the authors, along with the precomputed Structure-from-Motion (SfM) point clouds used for initialization. These scenes include \textit{Drjohnson} and \textit{Playroom} from Deep Blending \cite{deep-blending}, nine scenes from MipNeRF 360 \cite{barron2021mipnerf}, and the \textit{Train} and \textit{Truck} scenes from Tanks and Temples \cite{t&t}. 

We retained every eighth image from the provided input images as test images and computed Peak Signal-to-Noise Ratio (PSNR), Structural Similarity Index (SSIM), and Learned Perceptual Image Patch Similarity (LPIPS) \cite{lpips} for these test images.

Except for the adaptations discussed in Section \ref{sec:method}, we left all hyperparameters at the default values used in 3DGS. All scenes were trained with identical configurations to ensure consistency across evaluations.

\begin{figure*}[htbp]
\centering
\includegraphics[width=\textwidth]{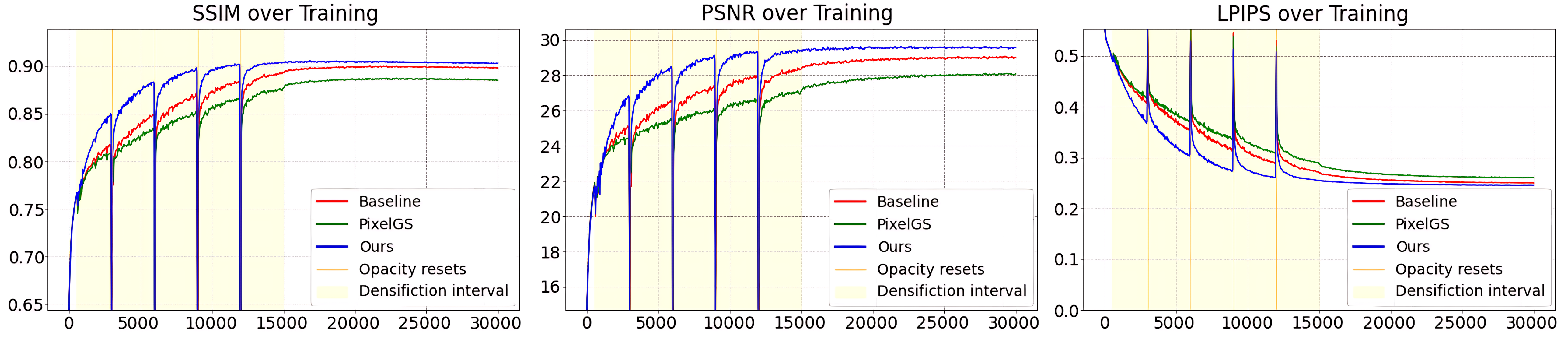}
\caption{Comparison of the metrics over all test images achieved by the baseline (3DGS) \cite{first_gs}, PixelGS \cite{pixelgs}, and Ours for the Drjohnson scene. PixelGS struggles with misplaced Gaussians, harming reconstruction quality.}
\label{fig:drjohnson_ours_upscled}
\end{figure*}

\subsection{Quantitative Evaluation}

As we directly benefit from the work of \cite{pixelgs} and \cite{error_based_dens}, incorporating implementations from their proposed methods, we evaluate our method against theirs. Table \ref{tab:combined_metrics} presents the evaluation metrics and the number of Gaussians used for scene reconstruction across \textit{3DGS}, \textit{PixelGS}, \textit{Revising Densification in Gaussian Splatting} and our method. To provide further context for our results, we have also included metrics for previous radiance field methods such as Plenoxels \cite{plenoxels}, INGP \cite{mueller2022instant}, and MipNeRF \cite{barron2021mipnerf}.

Our method consistently outperforms the baseline across all three datasets and evaluation metrics, using a comparable number of Gaussian primitives for scene representation. Moreover, we surpass \textit{PixelGS} and \textit{Revising Densification in Gaussian Splatting} in almost all metrics across all datasets. Notably, our method achieves these results using only about half the number of Gaussians required by PixelGS. 

In addition to the metrics reported in Table \ref{tab:combined_metrics}, we also highlight the rendering quality during training in Figure \ref{fig:drjohnson_ours_upscled}. Here, our method (blue) clearly outperforms 3DGS and PixelGS especially at the beginning of the training. After only 10k training iterations, the PSNR is already higher than the PSNR of the other methods after 30k iterations. After half of the training steps at 15k iterations, our method reaches a state where the quality does not improve much further. This suggests that with our method much shorter training times are possible without losing a lot of quality. To support this, we also report the rendering quality metrics after 15k iterations in Table \ref{tab:shorter_training}. Here, we observe considerably higher rendering quality when using our method. Furthermore, we also compare the quality between the models trained with 15k iterations to models trained with 30k iterations. Notably, our method already exceeds the final quality of 3DGS and PixelGS with only 15k training steps.

Training the scene with half of the iterations leads to a training time reduction in more than half of the time. This is because earlier training iterations take up less time since fewer Gaussians have to be rendered.

\begin{table}[htbp]
\centering
\caption{Rendering quality metrics when using 15k training iterations vs 30k iterations averaged over three scenes (one from each dataset). }
\label{tab:shorter_training}
\begin{tabular}{|l|rrr|}
\hline
	&   PSNR$\uparrow$	&	SSIM$\uparrow$	&	LPIPS$\downarrow$	\\
\hline
\textit{15K	Iterations} &&&\\	
3DGS	&	26.63	&	0.870	&	0.191	\\
PixelGS	&	26.33	&	0.869	&	0.183	\\
Ours	&	27.27	&	0.882	&	0.167	\\
\hline
\textit{30K	Iterations} &&&\\						
3DGS	&	27.23	&	0.878	&	0.174	\\
PixelGS	&	27.00	&	0.877	&	0.164	\\
Ours	&	\textbf{27.65}	&	\textbf{0.886}	&	\textbf{0.156}	\\
\hline
\end{tabular}
\end{table}

 \begin{figure*}[htbp]
\centering
\includegraphics[width=0.985\textwidth]{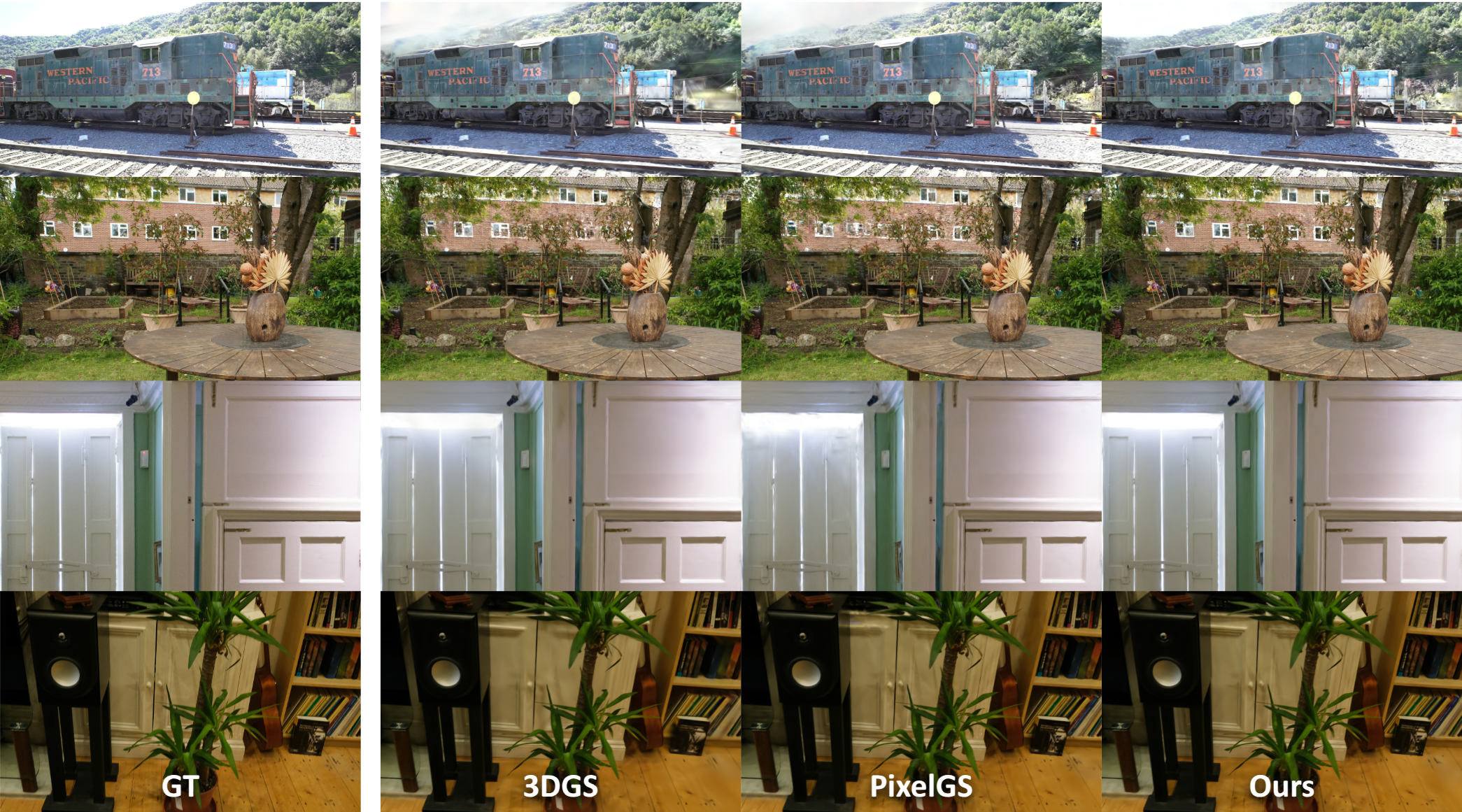}
\caption{Visual comparison between 3DGS, PixelGS and our method along with the corresponding ground truth. Our method shows less artifacts especially in background regions (first two rows) and less over-fitting noise (last two rows).}
\label{fig:comparison}
\end{figure*}

\subsection{Qualitative Evaluation}
To underline the quantitative results from Table \ref{tab:combined_metrics},
Figure \ref{fig:comparison} presents a selection of test views rendered using 3DGS, PixelGS and our method. As shown, our method consistently outperforms the other methods in terms of reconstruction quality. Our model excels at capturing high-complexity regions in both the foreground and background, where both 3DGS and PixelGS tend to populate these areas with large, blurry Gaussians. An example of improved background regions can be seen in the second row of Figure \ref{fig:comparison}. Here, especially the windows of the building in the background look much less distorted using our method. This is a similar result as shown in Figure \ref{fig:pruning_causes_trouble}, where the model was trained without pruning. A better reconstruction in the foreground regions can be seen in the last row. Focusing on the white door of the cupboard we observe blurry regions when using 3DGS and overfitted regions when using PixelGS. A similar observation can be made in the third row, looking at the door on the left. Furthermore, we often observe floating artifacts above the train in the first row, which our method resolves. Further examples can be viewed in the appendix.

\begin{table}[htbp]
\caption{Results for the ablation study, where we deactivate each of our three proposed components, as well as the pixel gradient from PixelGS \cite{pixelgs} and the opacity correnction from revising densification \cite{error_based_dens}. }
\label{tab:ablation_small}
\resizebox{\linewidth}{!}{
\begin{tabular}{|l|rrrr|}
\hline
Method / Metric &  PSNR$\uparrow$ & SSIM$\uparrow$ &  LPIPS$\downarrow$ &  \#Gaussians$\downarrow$ \\
\hline
Ours    &	\textbf{27.46}	&	0.841	&	\textbf{0.194} &	3,021k \\
\hline

w/o extent correction &  27.34&	0.841&	\textbf{0.194}&	3,204k\\
w/o pruning strategy &   27.37&	0.841&	\textbf{0.194}&	3,097k\\
w/o exp grad thresh  &	 27.45&	\textbf{0.842}&	0.195&	3,143k\\
\hline

w/o opacity correction&  27.30&	0.839&	0.191&	4,755k\\
w/o pixel gradient&      27.38&	0.834&	0.213&	\textbf{1,989k}\\
\hline

3DGS            &        27.16&	0.832&	0.216&	2,994k\\
\hline

\end{tabular}
}
\end{table}

\subsection{Ablation Study}

In total, our final method consists of five changed components compared to the default implementation of 3DGS \cite{first_gs}. Those include our three proposed methods, the pixel gradient from \cite{pixelgs} and the adapted opacity from \cite{error_based_dens}. To evaluate the effectiveness of our final configuration, we perform an ablation study in which we deactivate each of the components in an isolated experiment. The quantitative results for those experiments can be viewed in Table \ref{tab:ablation_small}. Here, we average all results from 13 scenes across all datasets. Generally, we observe that turning off any component directly leads to worse PSNR performance. The highest decrease across our methods is found with the correction of the \textit{scene extent} and the significance aware pruning strategy. The exponential \textit{gradient threshold}, on the other hand, does not show a big difference in PSNR. Nevertheless, deactivating the exponential \textit{gradient threshold} leads to more Gaussian splats being generated.

%% file: chapters/6_conclusion.tex
\newpage
\section{\uppercase{Conclusion}}
\label{sec:conclusion}

Gaussian Splatting has demonstrated its ability to surpass state-of-the-art reconstruction methods in quality; however, challenges such as under-reconstruction, artifacts, and the omission of important details, particularly in background regions highlight areas for improvement. These limitations often arise from imprecise densification. To address these limitations, we build on recent advances in Adaptive Density Control for 3DGS and propose several novel improvements: a correction mechanism for \textit{scene extent}, an exponentially ascending \textit{gradient threshold}, and significance-aware pruning.

Our comprehensive evaluation demonstrates that combining these techniques effectively addresses these challenges, resulting in improved reconstruction quality while maintaining a manageable number of Gaussian primitives. Although some of the modifications only bring minor improvements, all of the components are straightforward to implement into existing 3DGS frameworks, providing a practical and efficient enhancement to previous methods.


\section*{\uppercase{Acknowledgements}}
This work has partly been funded by the German Research Foundation (project 3DIL, grant no.~502864329), the German Federal Ministry of Education and Research (project VoluProf, grant no.~16SV8705), and the European Commission (Horizon Europe project Luminous, grant no.~101135724).

%% file: chapters/7_appendix.tex
\onecolumn
\section*{\uppercase{Appendix}}

\begin{figure*}[b!]
    \centering
    \includegraphics[width=1\textwidth]{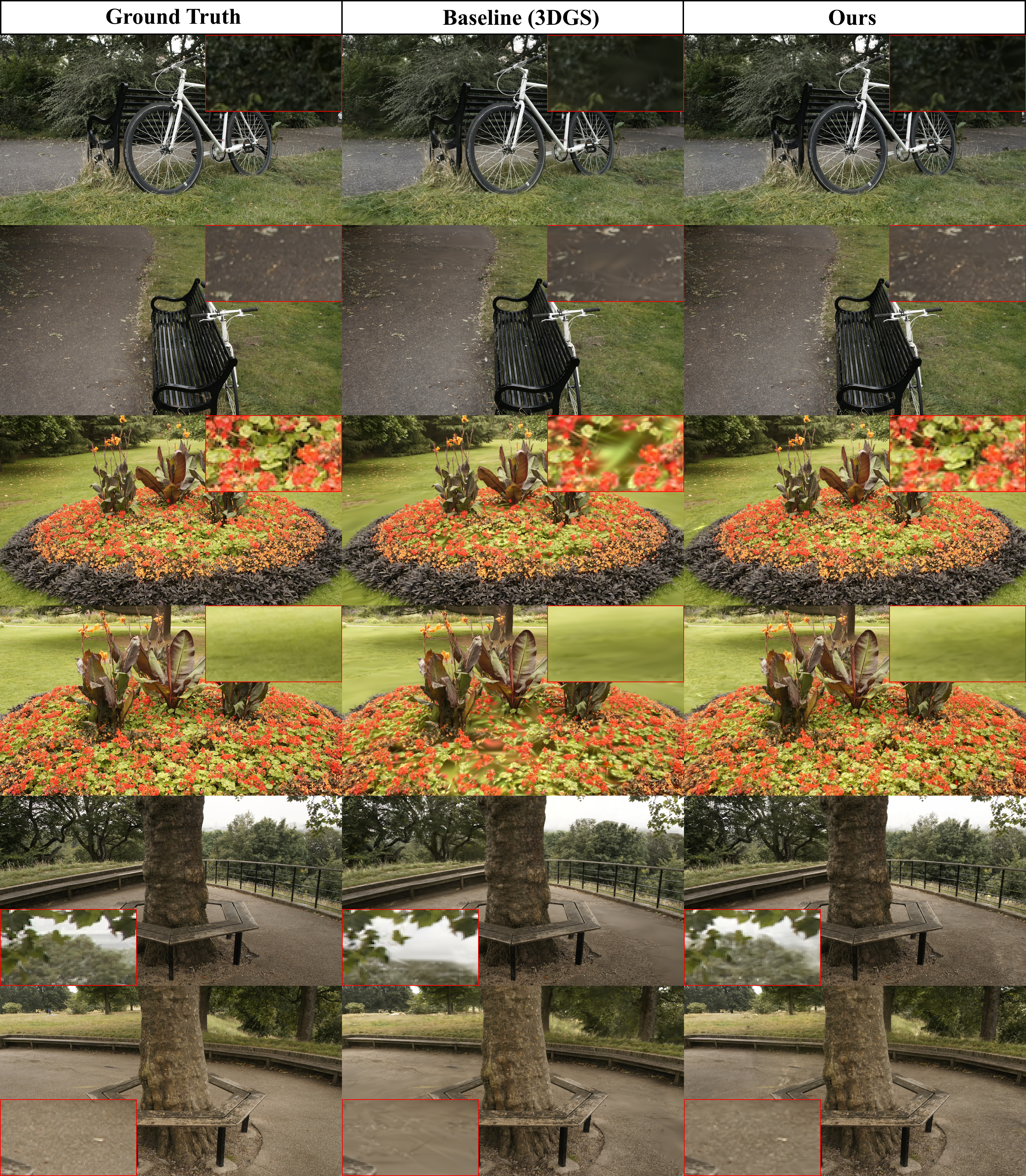}
    \caption{Rendered novel views for the Bicycle, Flowers, and Treehill scenes from the MipNeRF-360 dataset \cite{barron2021mipnerf} demonstrate that our model enables detailed reconstruction of intricate structures in both foreground and background regions. In comparison, our approach significantly surpasses the baseline in terms of reconstruction quality.}
    \label{fig:mipnerf}
\end{figure*}

\begin{figure*}
    \centering
    \includegraphics[width=1\textwidth]{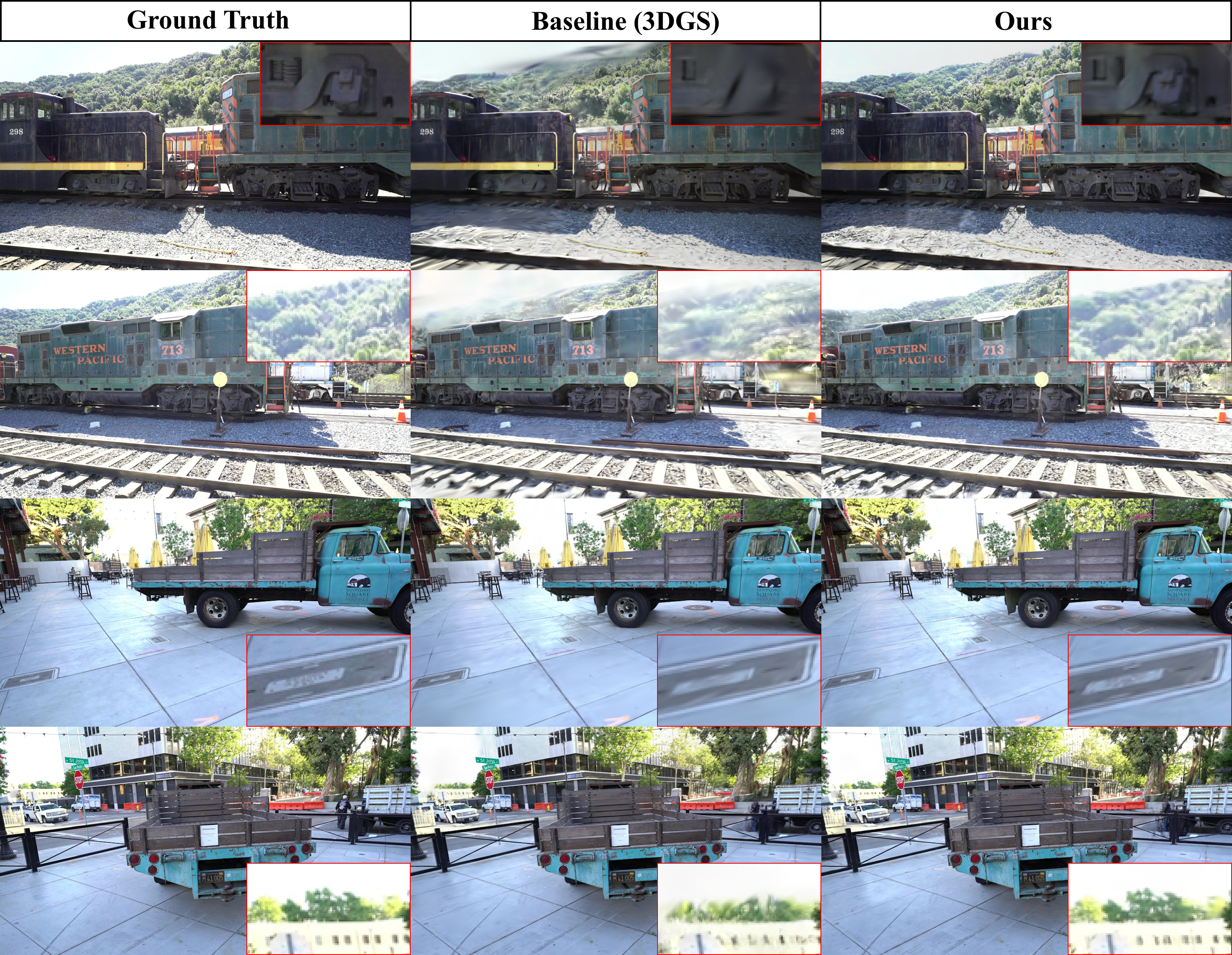}
    \caption{Rendered novel views for the Train and Truck scenes from the Tanks and Temples dataset \cite{t&t} clearly demonstrate that our model improves reconstruction, particularly in background regions and near scene edges, while reducing the number of artifacts. As a result, our approach significantly outperforms the baseline in terms of reconstruction quality.}
    \label{fig:t&t}
\end{figure*}